\documentclass[journal]{IEEEtran}
\usepackage{cite}
\usepackage{multirow}
\usepackage{amsmath}
\usepackage{graphicx}
\usepackage{booktabs}
\usepackage{subfigure}
\usepackage{stfloats}
\usepackage{makecell}

\usepackage{color}
\usepackage{amssymb}
\usepackage{algorithm}
\usepackage{algorithmic}
\usepackage{array}
\usepackage[table,xcdraw]{xcolor}

\begin{document}
%
\title{Spatial-Spectral Clustering with Anchor Graph for Hyperspectral Image}
%
%
%

\author{
        Qi~Wang,~\IEEEmembership{Senior Member,~IEEE,}
        Yanling~Miao,
        Mulin~Chen,
        and~Xuelong~Li,~\IEEEmembership{Fellow,~IEEE}
\thanks{ The authors are with the School of Computer Science and the Center for OPTical IMagery Analysis and Learning (OPTIMAL), Northwestern Polytechnical University, Xi'an 710072, China (e-mail:  crabwq@gmail.com, skilamiaomyl@gmail.com, chenmulin001@gmail.com, xuelong\_li@nwpu.edu.cn) (\textit{Corresponding author: Xuelong Li.})}}

\maketitle

\begin{abstract}
Hyperspectral image (HSI) clustering, which aims at dividing hyperspectral pixels into clusters, has drawn significant attention in practical applications. Recently, many graph-based clustering methods, which construct an adjacent graph to model the data relationship, have shown dominant performance. However, the high dimensionality of HSI data makes it hard to construct the pairwise adjacent graph. Besides, abundant spatial structures are often overlooked during the clustering procedure. In order to better handle the high dimensionality problem and preserve the spatial structures, this paper proposes a novel unsupervised approach called spatial-spectral clustering with anchor graph (SSCAG) for HSI data clustering. The SSCAG has the following contributions: 1) the anchor graph-based strategy is used to construct a tractable large graph for HSI data, which effectively exploits all data points and reduces the computational complexity; 2) a new similarity metric is presented to embed the spatial-spectral information into the combined adjacent graph, which can mine the intrinsic property structure of HSI data; 3) an effective neighbors assignment strategy is adopted in the optimization, which performs the singular value decomposition (SVD) on the adjacent graph to get solutions efficiently. Extensive experiments on three public HSI datasets show that the proposed SSCAG is competitive against the state-of-the-art approaches. 
\end{abstract}

\begin{IEEEkeywords}
Hyperspectral image, graph-based clustering, anchor graph, spatial-spectral information.
\end{IEEEkeywords}

\IEEEpeerreviewmaketitle

\section{Introduction}

\IEEEPARstart{H}{yperspectral} image (HSI) data obtained by hyperspectral imaging spectrometer provides abundant spatial structure and spectral information of the observed objects~\cite{xu2015intrinsic, wang2020afast}. With very narrow diagnostic spectral bands (wavelength interval is generally 10 $ nm $), HSI usually has high spectral resolution~\cite{deng2018active}, and can effectively distinguish subtle objects and materials between land cover classes~\cite{8052584, mei2018simultaneous}. Therefore, HSI has been applied in the real-world applications like e.g. vegetation investigation~\cite{9030286}, resource exploration~\cite{8075394}, environmental monitoring~\cite{8898447} and target identification~\cite{DBLP:journals/corr/abs-1905-01662}. Among current tasks, the clustering is a commonly used method for processing hyperspectral image. The purpose of HSI clustering is to partition all pixels into several groups according to their intrinsic properties. By assigning groups, the points in one group have high similarity, while those in different groups show great differences.

Though numerous techniques have been developed, HSI clustering remains to be a challenging issue. Traditional approaches include the spectral clustering~\cite{ng2002spectral}, spectral curvature clustering~\cite{chen2009spectral}, the multiview clustering~\cite{2017A}, the subspace learning~\cite{7425209, Li2017Locality, article} etc. These methods usually have high time complexity which comes from two major parts: (1) the construction of $ n\times n $ adjacent graph takes $ {\cal O}(n^{2}d) $, where $ n $ and $ d $ are the number of pixels and spectral bands respectively; (2) the eigenvalue decomposition on the graph Laplacian matrix costs $ {\cal O}(n^{2}c) $ or $ {\cal O}(n^{3}) $, where $ c $ is the number of land cover types. To address the above problems, some anchor graph-based approaches have been proposed. The Anchor Graph (AG) algorithm~\cite{liu2010large} utilizes a small number of anchors enough to cover the whole points to construct the large-scale adjacent graph. It reduces the computational complexity to $ {\cal O}(ndm) $, where $ m $ is the number of anchors. Motivated by this, various AG-based variants have been developed over the past years~\cite{he2020fast, wang2019scalable}. 

However, the AG-based methods ignore the spatial information within pixels, which limits their discriminant capability for real-world applications~\cite{huang2019dimensionality}. Therefore, many attempts have been made in spatial-spectral combined approaches through spatial correlations and spectral information. This category of methods mainly utilize the spatial correlations in the filtering preprocessing, which filter local homogeneous regions with single scale~\cite{zhou2014dimension, feng2014discriminative}. Nevertheless, they cannot fully capture the spatial structures for two reasons: (1) the HSI data may include small and large homogeneous regions simultaneously, but the local homogeneous regions can not be covered accurately with the single-scale filter. This phenomenon makes AG-based clustering algorithms fail to describe different spatial structures of HSI; (2) most of them just consider the spatial structures in the HSI data preprocessing process, and fail to incorporate these inherent structures into the clustering process.

To overcome the aforesaid shortcomings, this paper proposes a new spatial-spectral clustering with anchor graph (SSCAG) for HSI data. The main contributions can be summarized as follows. 

$\bullet$ We adopt the anchor graph-based strategy to construct the adjacent graph, which encodes the similarity between the data points and the anchors. Benefited from the anchor graph, the data structure is exploited effectively and the computational complexity is reduced. 

$\bullet$ We introduce a new distance metric that skillfully combines spatial structure and spectral features, which is able to select representative neighbors. In this way, the spatial-spectral collaboration information are integrated into the proposed model.

$\bullet$ We design an effective neighbors assignment strategy with spatial structures to learn the adjacent graph. Different from previous works, the proposed model employs SVD to get the final results, which is more efficient than eigenvalue decomposition.

The reminder of this paper is organized as follows. Section II introduces an overview of traditional and AG-based HSI clustering methods. Section III describes the proposed SSCAG in detail. Section IV presents the extensive experimental results and corresponding analysis. Finally, the conclusion is summarized in Section V. 

\begin{figure*}[htbp]
	\centering
	\includegraphics[height=4.67cm,width=18cm]{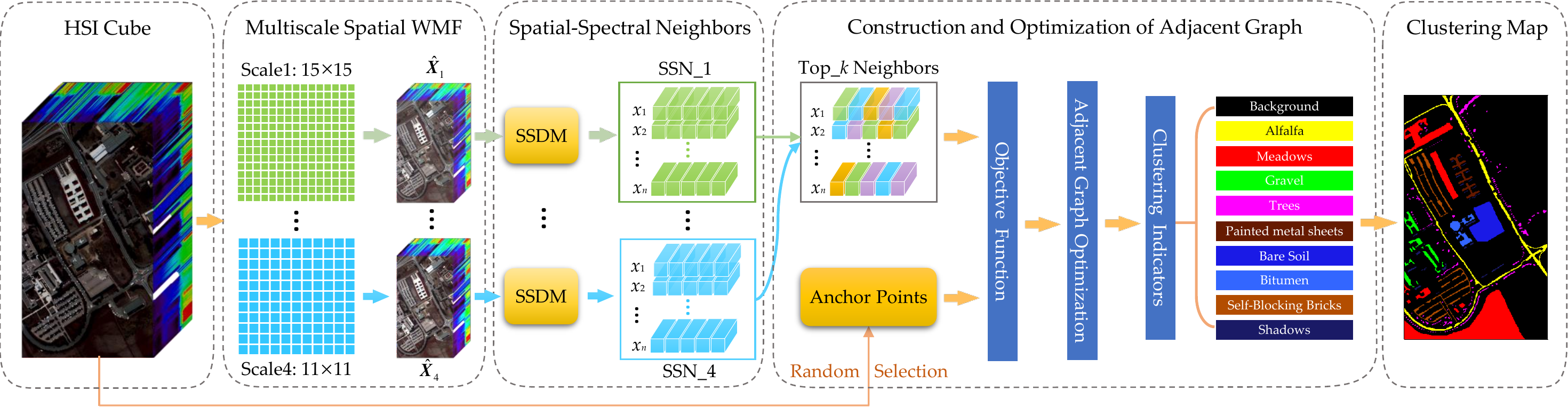}
	\caption{Overview architecture of the proposed SSCAG. The multiscale spatial WMF first is used to process HSI data. Then the spatial-spectral neighbors of each pixel are acquired by SSDM. Next we utilize anchor points and $ k $ nearest spatial-spectral neighbors to construct objective function. By adopting an effective neighbors assignment strategy with spatial information, the final clustering is obtained. \textbf{SSDM}: spatial-spectral distance metric. \textbf{SSN\_a}: spatial-spectral neighbors set. }
	\label{fig:overview}
\end{figure*} 

\section{Related Work}

In this section, we first briefly review the related works on traditional clustering methods, and then some existing AG-based clustering methods are introduced.
 
\subsection{Traditional Clustering Methods}

Many fundamental researches on HSI data clustering have been proposed. The traditional methods contains centroid-based approaches~\cite{hartigan1979algorithm, bezdek1984fcm, chen2004robust}, density-based approaches~\cite{rodriguez2014clustering, 2011Efficient} and graph-based approaches~\cite{2016Spectral, 2018A}.

The centroid-based approaches usually cluster HSI data based on the similarity measure. Ren \textit{et al.}~\cite{8899113} proposed an improved $ k $-means clustering to identify the mineral types from HSI of mining area, which uses dimensionless similarity measurement methods to obtain the mapping results, enhancing spectral absorption features. Chen \textit{et al.}~\cite{chen2004robust} proposed a fuzzy c-means (FCM) clustering with spatial constraints, which contributes to the introduction of fuzziness for belongingness of each pixel and exploits the spatial contextual information. Salem \textit{et al.}~\cite{inproceedings} developed hyperspectral feature selection for HSI clustering, which utilizes revisited FCM with spatial and spectral features to enhance the clustering. 
The density-based approaches form clusters by dense regions in the feature space. Tu \textit{et al.}~\cite{8732685} applied density peak clustering for noisy label detection of HSI, which considers the spatial correlations of adjacent pixels to define the local densities of the training set, improving the performance of classifiers.
The graph-based methods find clusters by learning the adjacent graph of data. Zhang \textit{et al.}~\cite{2016Spectral} took the spectral correlations and spatial information into sparse subspace clustering (SSC), which obtains a more accurate coefficient matrix for building the adjacent graph and promotes the clustering performance. To overcome the single sparse representation in SSC model, Yan \textit{et al.}~\cite{2018A} proposed two adjacent graph based on overall sparse representation vector and dynamic weights selection method, which better uses the spectral relationship and spatial structure during adjacent graph construction.

As mentioned above, there are some drawbacks about them. The centroid-based approaches are sensitive to initialization and noise. The density-based approaches are not suitable for HSI data, because it is hard to find density peaks in high-dimensional sparse feature space. The traditional graph-based is inefficient in large scale data, which has high computational complexity.

\subsection{AG-based Clustering Methods}
AG-based methods can address the scalability issue via a small number of anchors which adequately cover the entire points. Motivated by recent development in the AG construction, He $ et $ $ al. $~\cite{he2020fast} proposed fast semi-supervised learning with AG for HSI data, which constructs a naturally sparse and scale invariant AG, alleviating the computation burden. Wang $ et $ $ al. $~\cite{wang2019scalable} developed a scalable AG-based clustering method, which adds the non-negative relaxation to AG model. With this, the clustering results  are directly obtained without adopting $ k $-means. To handle the defect that AG-based methods usually ignore the spatial information, Wang $ et $ $ al. $~\cite{wang2017fast} presented fast spectral clustering with AG for HSI data, which fuses the spatial information by using the mean of neighboring pixels to reconstruct center pixel. Wei $ et $ $ al. $~\cite{wei2019fast} utilized the spatial correlation by adopting weighted mean filtering (WMF) to filter hyperspectral pixels, which considers the local neighborhood relationship within a window. Zhou $ et $ $ al. $~\cite{zhou2014dimension} proposed a regularized local discriminant embedding model with spatial-spectral information, which incorporates spatial feature into the dimensionality reduction procedure, to achieve optimal discriminative matrix by the minimization of local spatial–spectral scatter. Feng $ et $ $ al. $~\cite{feng2014discriminative} designed the discriminate margins with spatial-spectral neighborhood pixels, which 
effectively captures the discriminative features and learns the structures of HSI data. Luo $ et $ $ al.$~\cite{luo2018feature} constructed the intraclass spatial-spectral hypergraph by considering the coordinate relationship and similarity between adjacent samples, which can better make the high-dimensional spatial features embedded in low-dimensional space.

\section{Spatial-Spectral Clustering \\ with Anchor Graph}
\label{SSCAG}
In this section, we detail the proposed SSCAG algorithm from four aspects: weighted mean filtering (WMF), spatial-spectral neighbors, anchor-based adjacent graph construction and anchor graph-based spectral analysis. The overview architecture of SSCAG is shown in Fig. \ref{fig:overview}.

\subsection{WMF}
To smooth the homogeneous regions and reduce the interference of noisy points in the HSI, the WMF is employed to preprocess the pixels by utilizing spatial information. Suppose that each pixel of HSI can be denoted as a vector $ x_{i}\in\mathbb{R}^{d} $ $ (i=1,2,...,n) $, where $ d $ is the number of spectral bands and $ n $ refers to the number of HSI pixels. The HSI data is denoted as $ \mathbf{X}=[x_{1},x_{2},...,x_{n}]^{T}\in \mathbb{R}^{n\times d} $, and $ s_{ij} $ represents the similarity between $ x_i $ and $ x_j $, $ \mathbf{S} \in \mathbb{R}^{n \times n} $. Let $ \ell(x_{i})\in\{1,2,...,c\} $ be the class label of the pixels $ \{x_{1},x_{2},...,x_{n}\} $ respectively, where $ c $ is the number of classes. Assuming that the coordinate of pixel $ x_{i} $ is denoted as $ (p_{i},q_{i}) $, the adjacent pixels centered at $ x_{i} $ can be defined as
\begin{equation}\label{Eq_1}
\Omega(x_{i})=\bigg\{ x_{i}(p, q) |
\begin{array}{rcl}
p\in [p_{i}-t,p_{i}+t]\\
q\in [q_{i}-t,q_{i}+t] 
\end{array}\bigg\},
\end{equation}
where $t=(w-1)/2$, $ w $ indicates the size of neighborhood window and is a positive odd number. The pixels in the neighborhood space $ \Omega(x_{i}) $ are also represented as $ \{x_{i1},x_{i2},...,x_{i(w^{2}-1)}\}$, where $ w^{2}-1 $ is the number of neighbors of $ x_{i} $.

The reconstructed pixel $ \hat{x}_{i} $ by WMF is defined with a weighted summation, i.e.,
\begin{equation}\label{Eq_2}
\hat{x}_{i}=\frac{ x_{i} +\begin{matrix} \sum_{k=1}^{w^{2}-1} v_{k}x_{ik} \end{matrix}}
{1+\begin{matrix} \sum_{k=1}^{w^{2}-1} v_{k} \end{matrix}}, \quad x_{ik}\in \Omega(x_{i}),
\end{equation}
where $ v_{k}=exp\{-\gamma_{0}\|x_{i}-x_{ik}\|_2^{2}\} $ is the weight that represents the spectral similarity between $ x_{ik} $ and $ x_{i} $. The parameter $ \gamma_{0} $ is empirically set to
be 0.2 in the experiments, which reflects the degree of filtering. After filtering, the consistency of pixels in the same homogeneous regions is guaranteed. Since HSI may include homogeneous regions of different sizes simultaneously, we use multiscale WMF to obtain potential spatial structures of HSI. The abundant information with multiscale complementarity is beneficial to enhance the description for local homogeneous regions, thereby improving the performance of clustering.

\subsection{Spatial-Spectral Neighbors}
\label{ssn}
To better explore the comprehensive characteristics of HSI data, we introduce a new distance metric to seek effective neighbors by incorporating the spatial structures and spectral features. 

The pixels in the HSI are spatially related~\cite{8447427}. Especially, the adjacent pixels in a local homogeneous area have the spatial distribution consistency of land objects, which consist of the same materials and belong to the same category~\cite{zhang2016spectral}. Therefore, neighboring pixels are used to measure the spatial and spectral similarity. Suppose that a pixel $ x_{i} $ and its neighbors in $ \Omega(x_{i}) $ form a local pixel patch $ P(x_i)=\{x_{i1}, x_{i2}, ..., x_{iw^{2}}\} $. Let the spatial feature matrix be $ \mathbf{\hat{L}}=[l_1, l_2, ..., l_n] \in \mathbb{R}^{2 \times n} $, where $ l_i $ denotes the coordinate of $ x_i $. The distance $ d_{ss} $ of $ x_{i} $ and $ x_{j} $ can be defined as
\begin{equation}\label{Eq_3}
d_{ss}(x_{i},x_{j})=\frac{\begin{matrix}\sum_{h=1}^{w^{2}} v_{ih} \|x_{ih}-\hat{x}_{j}\|_2 \end{matrix}}
{\begin{matrix} \sum_{h=1}^{w^{2}} v_{ih} \end{matrix}}, \quad x_{ih} \in P(x_{i}),
\end{equation}
where $ v_{ih} $ is the weight that indicates the spatial similarity between pixel $ x_{ih} $ and $ \hat{x}_{j} $. The weight can be obtained by a kernel function, which is denoted as
\begin{equation}\label{Eq_4}
v_{ih}=exp\Big\{-\|l_{ih}-l_j \|_2^{2}/\sigma_j^2 \Big\},
\end{equation}
where $ l_{ih} $ and $ l_j $ are the coordinates of $ x_{ih} $ and $ x_j $, respectively.  The spatial distance $ \|l_{ih}-l_j \|_2 $ is defined by the Euclidean distance between their coordinates, and $ \sigma_j $ is set as the average of $ \begin{matrix} \sum_{h=1}^{w^2} \|l_{ih}-l_j \|_2  \end{matrix} $, i.e.,
\begin{equation}\label{Eq_5}
\sigma_j=\frac{1}{w^2} \sum_{h=1}^{w^2} \|l_{ih}-l_j \|_2.
\end{equation}

Note that the values of $ \mathbf{X} $ and $ \mathbf{\hat{L}} $ are normalized within [0,1]. Eq. (4) and Eq. (5) enforce the pixels with larger spatial distances to have a smaller similarity.

For HSI data, the spectral neighbors may contain the pixels with similar spectrum, but they are in different classes. The spatial neighbors only consider the coordinate distance of pixels, which may include pixels placed in different classes, especially at the boundary. According to Eq. (3), the proposed $ d_{ss}(x_{i}, x_{j}) $ combines the spatial and spectral features simultaneously, where $\|x_{ih}-\hat{x}_{j}\|_2 $ denotes the spectral similarity of two pixels, and $ v_{ih} $ is the corresponding spatial similarity. Furthermore, it not only considers the neighboring pixels in the patch $ P(x_{i}) $, but also employs the adjacent pixels in the $ \Omega(x_{j}) $. Hence, $ d_{ss} $ chooses the effective neighbors by collaborating spatial and spectral distance. After obtaining the spatial-spectral neighbors set of each pixel with different scales of WMF, we choose $ k $ nearest neighbors for constructing the objective function in the next part.

\subsection{Anchor-based Adjacent Graph Construction}
To reduce the computational complexity of constructing adjacent graph, we exploit an anchor-based strategy to learn the adjacent graph, and design an efficient strategy to accelerate the optimization. Similar to the previous works~\cite{deng2013visual, deng2014weakly, cai2014large, nie2017unsupervised, nie2020unsupervised, li2019adaptive}, the anchor-based strategy mainly consists of two steps: 1) anchors generation; 2) adjacent graph construction.  

\textbf{Anchors Generation:} in large-scale clustering, the anchors are usually generated by $k$-means method or random sampling. We employ the random selection method to generate $ m $ anchors ($m\ll n$), since its computational complexity is $ {\cal O}(1) $. Let $ \mathbf{U} =[u_1,u_2,...u_m]^T\in \mathbb{R}^{m\times d} $ denotes the anchor set, and $ u_{\phi_i} $ represents the set of $ k $-nearest anchors for $ x_i $.

\textbf{Adjacent Graph Learning:} let $ \mathbf{Z} $ $\in \mathbb{R}^{n\times m}$ be the adjacent graph, where $ z_{ij} $ denotes the similarity between $ x_i $ and $ u_j $. $ \mathbf{Z}$ is constructed by $ k $-nearest neighbors method. Traditional approaches usually adopt kernel-based strategy to assign neighbors~\cite{hastie2009elements}, which always bring extra hyperparameters. Inspired by~\cite{chen2004robust} and~\cite{nie2016constrained}, we design an effective neighbor assignment strategy. The nearest anchors assignment of $ x_i $ can be seen as solving the objective function: 
\begin{equation}\label{Eq_6}
\min_{z_i^T \textbf{1}=1, z_{ij}\ge 0} 
\sum_{j=1}^{m} \underbrace {\|x_i-u_j\|_2^2 z_{ij}}_{\mathcal{J}_1} + \underbrace {\alpha \|\widetilde{x}_i-u_j\|_2^2 z_{ij} }_{\mathcal{J}_2} + \underbrace {\gamma z_{ij}^2}_{\mathcal{J}_3},
\end{equation}
where $ z_i^T $ is the \textit{i}-th row of $ \mathbf{Z} $, and $ \|x_i-u_j\|_2^2 $ is the square of Euclidean distance between pixel $ x_i $ and anchor $ u_j $. $ \widetilde{x}_i $ denotes the average of $ k $ spatial-spectral neighbors about $ x_i $, which can be calculated in section \ref{ssn}. Similarly, $ \|\widetilde{x}_i-u_j\|_2^2 $ is the distance between $ \widetilde{x}_i $ and $ u_j $. It can be seen that the objective function (6) contains three terms: $ \mathcal{J}_1 $, $ \mathcal{J}_2 $ and $ \mathcal{J}_3 $. $ \mathcal{J}_1 $ utilizes the spectral features to learn $ \mathbf{Z} $; $ \mathcal{J}_2 $ incorporates the local spatial-spectral feature to learn $ \mathbf{Z} $, where $\alpha$ controls the tradeoff between $ \mathcal{J}_1 $ and $ \mathcal{J}_2 $; $ \mathcal{J}_3 $ is the regularization term, which is to prevent the trivial solution of Eq. (6), where $ \gamma $ is the regularization parameter.

Let us define
\begin{equation}\label{Eq_7}
e_{ij}=\|x_i-u_j\|_2^2,
\end{equation}
\begin{equation}\label{Eq_8}
\widetilde{e}_{ij}=\|\widetilde{x}_i-u_j\|_2^2,
\end{equation}
where $ e_i \in \mathbb{R}^{m \times 1} $ denotes a vector with the \textit{j}-th element as $ e_{ij} $. Similarly, $ \widetilde{e}_i \in \mathbb{R}^{m \times 1 } $ is  a vector with the \textit{j}-th element as $ \widetilde{e}_{ij} $. Combining Eq. (7) and Eq. (8), we obtain
\begin{equation}\label{Eq_9}
E_{ij}=e_{ij}+\alpha \widetilde{e}_{ij},
\end{equation}
and denote $ E_i \in \mathbb{R}^{m \times 1} $ as a vector with the \textit{j}-th element as $ E_{ij} $, then the objective function (6) can be expressed as
\begin{equation}\label{Eq_10}
\min_{z_i^T \textbf{1}=1, z_{ij}\ge 0}  \frac{1}{2} \Big \|z_i+\frac{1}{2 \gamma}E_i \Big \|_2^2.
\end{equation}

The Lagrangian function of Eq. (10) is
\begin{equation}\label{Eq_11}
\mathcal{L}(z_i, \eta, \beta_i)=\frac{1}{2} \Big \|z_i+\frac{E_i}{2 \gamma} \Big \|_2^2 - \eta(z_i^T \textbf{1}-1) - \beta_i^T z_i,
\end{equation}
where $ \eta $ and $ \beta_i^T \ge 0 $ are the Lagrangian multipliers. To achieve the optimal $ z_i^* $, it should satisfy that the derivative of Eq. (11) with respect to $ z_i^* $ is equal to zero, i.e.,
\begin{equation}\label{Eq_12}
z_i^*+ \frac{E_i}{2 \gamma}-\eta \textbf{1} -\beta_i=\textbf{0}.
\end{equation}

Then the \textit{j}-th element of $ z_i^* $ is
\begin{equation}\label{Eq_13}
z_{ij}^*+ \frac{E_{ij}}{2 \gamma}-\eta \textbf{1} -\beta_{ij}=0.
\end{equation}

By the KKT condition and constraints, to achieve the optimal solution $ z_i^* $ to the function (6) that has exactly $ k $ nonzero values, the $ \eta $ and $ \gamma $ are
\begin{align}\label{Eq_14}
&\eta = \frac{1}{k} + \frac{1}{2k\gamma}\sum_{j=1}^{k}E_{ij},\\ \label{Eq_15}
&\gamma=\frac{k}{2}E_{i,k+1}-\frac{1}{2} \sum_{j=1}^{k}E_{ij}.
\end{align}

Therefore, the optimal solution $ z_{ij}^* $ is as follows:
\begin{equation}\label{Eq_16}
z_{ij}^*=\frac{E_{i,k+1}-E_{ij}}
{kE_{i,k+1}- \begin{matrix} \sum_{j'=1}^{k} E_{ij'} \end{matrix}}.
\end{equation}

For detail deviation, please refer to \cite{nie2016constrained}. The calculation of $ z_{ij}^* $ only involves the basic operations: addition, subtraction, multiplication and division, which ensures the efficiency of the proposed method. In addition, there are no hyperparameters which may affect the stability of model in Eq.~(16). Then, we propose to normalize $\mathbf{Z}$ to be a  doubly stochastic matrix.

\textbf{Doubly Stochastic Graph Construction:} after getting the adjacent graph $ \mathbf{Z} $, the normalized adjacent graph $ \mathbf{S} $ can be computed as:
\begin{equation}\label{Eq_17}
\textbf{S} = \textbf{Z} \boldsymbol{\Lambda}^{-1} \textbf{Z}^T,
\end{equation}
where $ \boldsymbol{\Lambda} $ is a diagonal matrix whose \textit{j}-th element is represented as $ \Lambda_{jj}=\begin{matrix} \sum_{i=1}^{n} z_{ij} \end{matrix} $, and $ \boldsymbol{\Lambda} \in \mathbb{R}^{m \times m} $. Intuitively, the element $ s_{ij} $ of matrix $ \mathbf{S} $ is expressed as $ s_{ij}= z_i^T \Lambda^{-1} z_j $ that satisfies $ s_{ij}=s_{ji} $. Moreover, with $ z_{ij} \ge 0 $ and $ z_i^T \textbf{1}=1 $, it can be proved that matrix $ \mathbf{S} $ is positive semidefinite and doubly stochastic~\cite{liu2010large}. The above property of $\mathbf{S}$ is crucial for reducing the computation cost of optimization, and the details will be given later.

\subsection{Anchor Graph-based Spectral Analysis}
The objective function of spectral clustering is 
\begin{equation}\label{Eq_18}
\mathop {\min }\limits_{{{\bf{F}}^T}{\bf{F}} = {\bf{I}}} {\mathop{\rm Tr}\nolimits} ({{\bf{F}}^T}{\bf{LF}}),
\end{equation}   
where $ \mathbf{F} \in \mathbb{R}^{n \times c} $ denotes clustering indicators matrix. The Laplacian matrix is $\mathbf{L=D-S}$, where the \textit{i}-th element of the degree matrix $ \mathbf{D} $ is denoted as $ d_{ii}=\begin{matrix} \sum_{j=1}^{n} \end{matrix} s_{ij} $. The optimal solution of $ \mathbf{L} $ is obtained by enforcing eigenvalues decomposition on $ \mathbf{L} $, which is comprised of eigenvectors corresponding to the smallest $ c $ eigenvalues of $ \mathbf{L} $. However, 
this solution process requires $ {\cal O}(n^2c) $, which is not conducive to HSI clustering. 
According to the natural characteristics of $ \mathbf{S}$, we have
\begin{equation}\label{Eq_19}
d_{ii}= \sum_{j=1}^{n} s_{ij} = \sum_{j=1}^{n} z_i^T \Lambda^{-1} z_j = z_i^T \sum_{j=1}^{n} \Lambda^{-1} z_j = z_i^T \textbf{1} = 1.
\end{equation}

Therefore, the degree matrix $\mathbf {D}$ equals to the identity matrix $ \mathbf{I}$ exactly. Moreover, it can be seen that the matrix $ \mathbf{S} $ is automatically normalized. Considering that $ \mathbf{L=D-S=I-S} $, Eq. (18) is equivalent to solving this problem:
\begin{equation}\label{Eq_20}
\mathop {\max }\limits_{{{\bf{F}}^T}{\bf{F}} = {\bf{I}}} {\mathop{\rm Tr}\nolimits} ({{\bf{F}}^T}{\bf{SF}}).
\end{equation}  

Note that $ \mathbf{S} $ can also be denoted as 
\begin{equation}\label{Eq_21}
\mathbf{S} = \mathbf{A} \mathbf{A}^{T},
\end{equation}
where $ \mathbf{A}=\mathbf{Z} \Lambda^{-1/2} $, then we apply SVD on $ \mathbf{A} $ as follows:
\begin{equation}\label{Eq_22}
\mathbf{A} = \mathbf{U} \boldsymbol{\Sigma} \mathbf{V}^{T},
\end{equation}
where $ \mathbf{U}\in \mathbb{R}^{n\times n} $, $ \boldsymbol{\Sigma}\in \mathbb{R}^{n\times m} $ and $ \mathbf{V}\in \mathbb{R}^{m\times m} $. With respect to the property of SVD, there are $ \mathbf{U}^{T}\mathbf{U} = \mathbf{I} $ and  $ \mathbf{V}^{T}\mathbf{V} = \mathbf{I} $. So $ \mathbf{S} = \mathbf{A} \mathbf{A}^{T} = \mathbf{U} \boldsymbol{\Sigma} \mathbf{V}^{T} \mathbf{V}\boldsymbol{\Sigma}^{T} \mathbf{U}^{T}= \mathbf{U} \boldsymbol{\Sigma} \boldsymbol{\Sigma}^{T} \mathbf{U}^{T} = \mathbf{U} \boldsymbol{\Sigma}^{2} \mathbf{U}^{T} $. It indicates that the column vectors of $ \mathbf{U} $ are the eigenvectors of $ \mathbf{S} $. Instead of using eigenvalue decomposition on $ \mathbf{S}$, we adopt SVD on $ \mathbf{A}$ to get the relaxed continuous solution of $ \mathbf{F} $, which only costs $ {\cal O}(nmc+m^{2}c) $. Then, we perform $ k $-means method on $ \mathbf{F} $ to get the discrete solution.
\renewcommand{\algorithmicrequire}{\textbf{Input:}}
\renewcommand{\algorithmicensure}{\textbf{Output:}}
\begin{algorithm}
	\caption{SSCAG}
	\label{Algorithm 1}
	\small
	\begin{algorithmic}[1]
		\REQUIRE HSI data $ \mathbf{X} \in \mathbb{R}^{n \times d} $, class number $ c $, window size $ w $, anchor number $ m $, neighbor number $ k $, the balance parameter $ \alpha $ and $ \gamma_{0}=0.2 $.
		\FOR {  $ i=1 $ \textbf{to}  $ n $ } 
		\STATE seek the neighbors of $ x_i $ according to Eq. (1);
		\FOR { $ k=1 $ \textbf{to} $ w^2-1 $  } 
		\STATE calculate the weights: $ v_{k}=exp\{-\gamma_{0}\|x_{i}-x_{ik}\|^{2}\} $; 
		\ENDFOR
		\STATE receive the filtered point $ \hat{x}_i $ by Eq. (2);
		\ENDFOR
		\STATE Obtain $ \mathbf {\hat{X}}=[\hat{x}_1,\hat{x}_2,...,\hat{x}_i,...,\hat{x}_n] $;
		\FOR { $ i=1 $ \textbf{to}  $ n $ }
		\STATE calculate the spatial-spectral distance $ d_{ss} $ by Eq. (3);\STATE find $ k $ spatial-spectral nearest neighbors for $ x_i $, computing the average of neighbors and saving as $\widetilde{x}_{i}$;
		\ENDFOR
		\STATE Generate $ m $ anchors by random sampling, obtaining anchor matrix $ \mathbf{U} \in \mathbb{R}^{d\times m} $;
		\STATE Find $ k $ nearest neighbors of $ x_i $ in $ \mathbf{U} $, saving the index set $ \phi_i $;
		\STATE Define function $ \mathcal{L}(\textbf{\textit{z}}_i)=\|\textit{\textbf{x}}_i-u_{\phi_i}\|_2^{2} \textbf{\textit{z}}_i + \alpha \|\widetilde{\textit{\textbf{x}}}_{i}-u_{\phi_i}\|_2^{2} \textbf{\textit{z}}_i + \gamma \|\textbf{\textit{z}}_i\|_2^2 $;\\
		\STATE Obtain $ \gamma $ and the adjacent graph $ \mathbf{Z} $ by Eq. (16) and Eq. (17);
		\STATE Design the normalized adjacent graph $ \mathbf{S}= \mathbf{Z} \boldsymbol{\Lambda}^{-1} \mathbf{Z}^T $, $ \Lambda_{jj}=\begin{matrix} \sum_{i=1}^{n} z_{ij} \end{matrix} $;
		\STATE Obtain $ \mathbf{F} $ by applying SVD on matrix $ \mathbf{A} $, and $ \mathbf{A}=\mathbf{Z} \Lambda^{-1/2} $; 
		\STATE Utilizing $ k $-means on $ \mathbf{F} $ for final clustering results;  
		\ENSURE Label matrix about $ c $ classes
	\end{algorithmic}
\end{algorithm}

The detail steps of SSCAG are described in \textbf{Algorithm 1}. 
The computational complexity of \textbf{Algorithm 1} mainly comes from four parts:
1) the WMF-based spatial preprocessing takes $ {\cal O}(ndw^2) $; 
2) obtaining $ m $ anchors by random selection costs $ {\cal O}(1) $, and the spatial-spectral combined distance $ d_{ss} $ is calculated with $ {\cal O}(n^2w^2) $; 
3) the cost of constructing $ \mathbf{Z} $ is ${\cal O}(ndm)$, and achieving the relaxed continuous eigenvalue of $ \mathbf{F}$ needs ${\cal O}(nmc+m^2c)$; 
4) the $ k $-means which is used to get the final clustering results takes $ {\cal O} (nmc\tau) $, where $ \tau $ is the number of iterations. Note that $ m \ll n $, $ c \ll d $, $ w $ and $ \tau $ are small. Therefore, the final computational complexity of SSCAG is $ {\cal O}(n^2w^2+ndm) $.

\section{Experiments}
In this section, to demonstrate the effectiveness of the proposed SSCAG method, the verification experiments are conducted on public hyperspectral datasets. Some excellent algorithms are adopted as competitors. Finally, the experimental results and corresponding analysis are provided.
 
\subsection{Dataset Description}
The detailed information of three hyperspectral datasets are introduced as follows. Their false color images and ground truths are displayed in Fig. \ref{fig:dataset}.

\subsubsection{Indian Pines} The hyperspectral image was acquired via AVIRIS device (spatial resolution is 20 $ m $) over the northwestern Indiana in 1992, as shown in Fig. \ref{fig:dataset} (a). The spatial size of this area is 145$\times$145 pixels. Each pixel contains 220 spectral bands ranging from 0.4 to 2.5 $\mu m$. Due to noise and water absorption phenomena, 20 channels are removed, leaving 200 bands to be used for experiments. The dataset contains 16 land cover types, and the distribution of samples is clear in Fig. \ref{fig:dataset} (b).  
\subsubsection{Pavia University} The scene was captured by the German ROSIS sensor (spatial resolution is 1.3 $ m $) over the Pavia in 2002, as displayed in Fig. \ref{fig:dataset} (c). The HSI data contains 610$\times$340 pixels and 115 spectral bands, and the spectral range is 0.43 to 0.86 $\mu m$. With discarding 12 noise and water absorption bands, 103 bands are left for classification task. This dataset includes 9 ground-truth classes, and the distribution of them is shown in Fig. \ref{fig:dataset} (d). 
\subsubsection{Salinas} 
\begin{figure}[htbp]\centering
	\includegraphics[width=8cm,height=14cm]{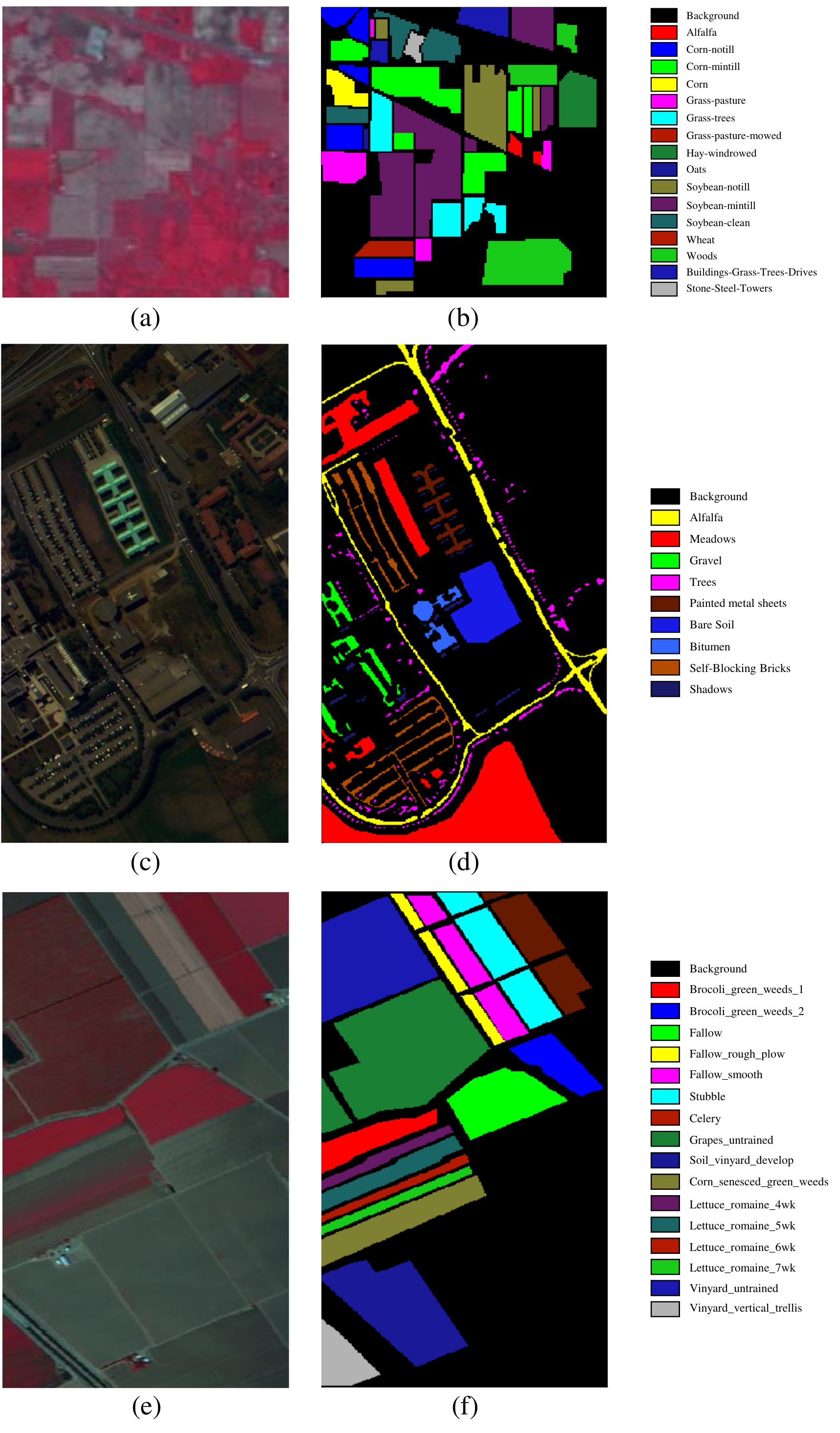}
	\caption{The RGB false color images and ground truth maps of three HSI datasets. (a) Indian Pines image of bands 50, 27 and 17. (c) Pavia University image of bands 60, 30 and 2. (e) Salinas image of bands 70, 27 and 17. (b), (d) and (f) are corresponding ground truth maps respectively.}\label{fig:dataset}
\end{figure}

Like the Indian Pines image, the Salinas data was also taken by AVIRIS sensor over the Salinas Valley in 1998, as exhibited in Fig. \ref{fig:dataset} (e). Unlike Indian Pines, its spatial resolution is 3.7 $ m $. Similarly, the 204 bands are left after excluding 20 water absorption bands. The size of Salinas image is 512$\times$217. All pixels in HSI data are divided into 16 categories whose distribution is illustrated in Fig. \ref{fig:dataset} (f).

\subsection{Experimental Setup}

\subsubsection{Comparison Algorithms} To validate the superiority of the proposed SSCAG, several HSI clustering methods are considered as follows.

a) \textit{Traditional clustering methods}: $ k $-means~\cite{hartigan1979algorithm}, FCM~\cite{bezdek1984fcm} and FCM-S1~\cite{chen2004robust}. These methods are typical centroid-based clustering approaches. $k$-means and FCM commonly use Euclidean distance as similarity measure to achieve the clusters. $k$-means is a hard clustering method which assigns each sample into a certain cluster. FCM belongs to soft clustering, which uses the membership to identify the relationship between samples and each cluster. FCM-S1 has enhanced robustness of the original clustering algorithms by exploiting spatial contextual information.  

b) \textit{AG-based clustering methods}: SGCNR~\cite{wang2019scalable}, FSCAG \cite{wang2017fast} and FSCS~\cite{wei2019fast}. The first method SGCNR only adopts the spectral feature, while other methods FSCAG and FSCS consider spatial and spectral information.  SGCNR builds an adjacent graph based on AG and the nonnegative relaxation conditions, which directly obtains the clustering indicators. FSCAG considers spectral correlation and spatial neighborhood properties of the HSI data to construct anchor graph. FSCS uses the spatial nearest pixels to reconstruct the center point within a window, which combines local spatial structure with spectral information.
   
\subsubsection{Evaluation Indices} Three quantitative metrics are used to evaluate clustering performance, including overall accuracy (OA), average accuracy (AA) and Kappa coefficient. OA represents the proportion of correctly classified pixels in HSI. AA denotes the average of the classification accuracy of each category. The value of OA and AA ranges from 0 to 1. Here, the higher accuracy value, the better clustering performance. Kappa coefficient is a metric combined commission error and omission error, which can evaluate the overall consistency. Its value range is [0,1]. The larger Kappa values indicate better consistency.

\begin{figure*}[htbp]
	\centering
	\includegraphics[width=18cm,height=8cm]{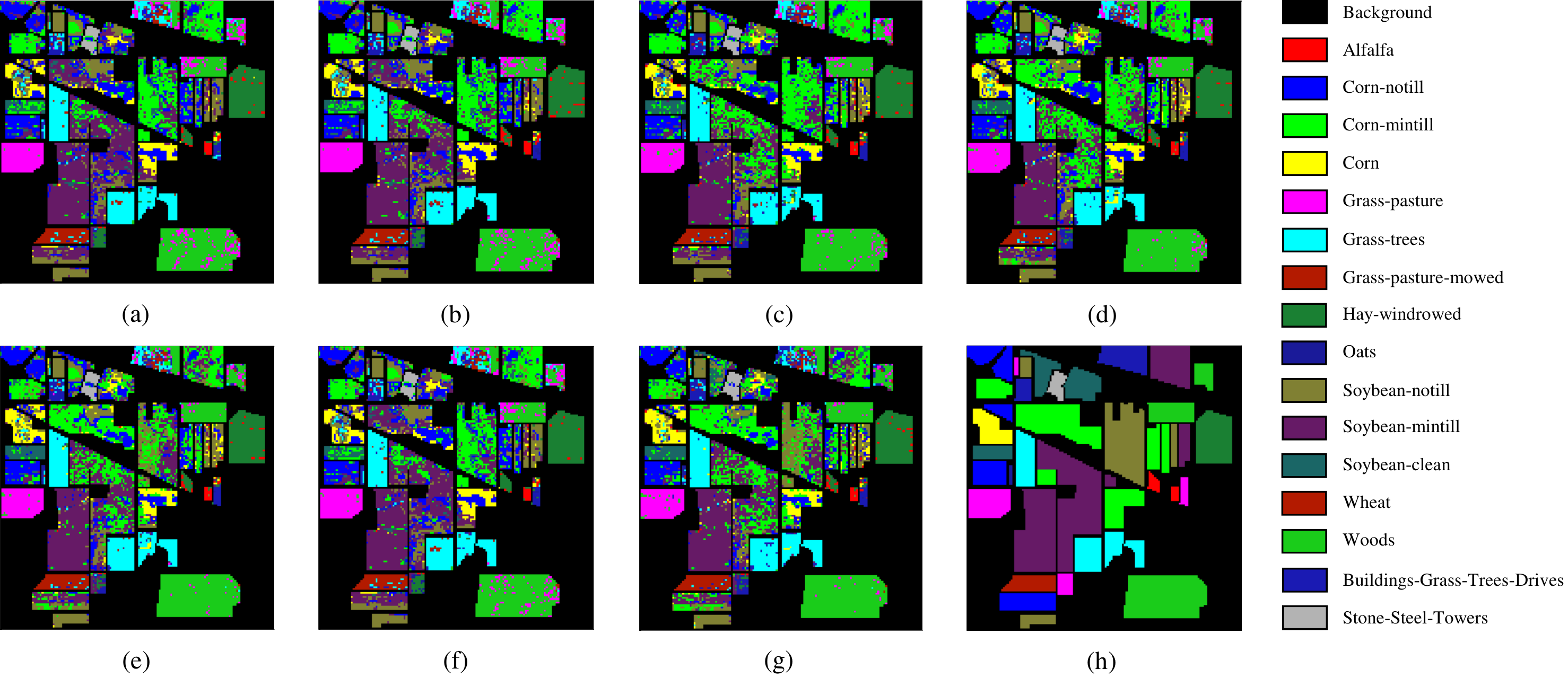}
	\caption{ Clustering maps and ground truth of Indian Pines. (a) $k$-means. (b) FCM. (c) FCM\_S1. (d) SGCNR. (e) FSCAG. (f) FSCS. (g) SSCAG. (h) Ground truth.}\label{fig:indian}
\end{figure*}

\begin{table*}[htbp] 
	\renewcommand{\arraystretch}{1.5}
	\caption{Quantitative metrics of different comparison methods and the proposed SSCAG on Indian pines dataset. The optimal value of each row is highlighted in bold.}
	\centering
	\label{table:indian}
	\centering
	\begin{tabular}{|p{3.6cm}<{\centering}||p{1.1cm}<{\centering}|p{1.1cm}<{\centering}|p{1.1cm}<{\centering}|p{1.1cm}<{\centering}|p{1.1cm}<{\centering}|p{1.1cm}<{\centering}|p{1.1cm}<{\centering}|}
		\hline
		Class     & $ k $-means& FCM & FCM\_S1 & SGCNR & FSCAG  & FSCS   & SSCAG \\ \hline
		\hline
		Alfalfa &0.0872	&\bf0.2449	&0.1252	&0.1547	&0.1143	&0.0187	    &0.1749     \\ \hline
		Corn-notill &0.2049	&0.2701	    &0.2334	&0.2765	&0.3397	&0.3245	    &\bf0.4097  \\ \hline
		Corn-mintill&0.1424	&0.1606	    &0.3404	&0.4131	&0.4190	&0.1467	    &\bf0.4431  \\ \hline  
		Corn	    &0.1612	&0.1689	    &0.1466	&0.2079	&0.3051	&0.2236	    &\bf0.5402  \\ \hline
		Grass-pasture&0.6067	&\bf0.6539	&0.6099	&0.5928	&0.5373	&0.5983	    &0.5039     \\ \hline
		Grass-trees	&0.7278	&0.7082	    &0.6988	&0.7124	&0.7987	&0.7837	    &\bf0.8395  \\ \hline
		Grass-pasture-mowed&0.8081	&\bf0.8215	&0.7985	&0.7548	&0.7665	&0.8004	    &0.7829     \\ \hline
		Hay-windrowed&0.8602	&0.7952	    &0.8727	&0.8825	&0.8573	&\bf0.9282	&0.8571     \\ \hline
		Oats     &0.3501	&0.4302	    &\bf0.4500&0.4459&0.3535&0.4315	    &0.3462     \\ \hline
		Soybean-notill	&0.1558	&0.1798	    &0.292	&0.2735	&0.3501	&0.2925	    &\bf0.4543  \\ \hline
		Soybean-mintill	&\bf0.6307&0.5381	&0.4792	&0.3934	&0.4038	&0.5605	    &0.5033     \\ \hline
		Soybean-clean	&0.0598	&0.1083	    &0.2124	&0.2753	&0.2590	&0.1192	    &\bf0.3531  \\ \hline
		Wheat	    &0.8086	&0.8107	    &0.8244	&0.8261	&0.8438	&0.8356	    &\bf0.8617  \\ \hline
		Woods	    &0.4012	&0.4145	    &0.4495	&0.4877	&0.7863	&0.4949	    &\bf0.8503  \\ \hline 
		Buildings-Grass-Trees-Drives&0.1288	    &0.1436	&\bf0.3054&0.2881&0.1716	&0.0736	&0.2877 \\ \hline
		Stone-Steel-Towers	&0.7161	&0.7017	    &0.7845	&0.8182	&0.8042	&0.7957	    &\bf0.8867  \\ \hline
		\hline
		\bf OA	&0.3929	&0.4128	&0.4551	&0.4531	&0.4708	&0.4409	&\bf0.5427 \\ \hline
		\bf AA	&0.4281	&0.4469	&0.4764	&0.4877	&0.5069	&0.4642	&\bf0.5684 \\ \hline
		\bf Kappa &0.3241	&0.3512	&0.3709	&0.3931	&0.4019	&0.3816	&\bf0.4818 \\ \hline	
	\end{tabular}
\end{table*}

\subsection{Experimental Results and Analyses}

\subsubsection{Parameter Settings} The whole pixels of the Indian Pines, Pavia University and Salinas were used as testing data. Set the number of clusters equal to the number of the ground truth labels in each dataset. For centroid-based clustering methods, all the parameters are manually adjusted to the optimum. Every algorithm is repeated 20 times to avoid the bias. For AG-based methods, the optimal parameter settings of SGCNR is followed as~\cite{wang2019scalable}. For FSCAG, the anchors number $ m $, neighbors number $ k $ and balance parameter $ \alpha $ are $ \{ m=500, k=5, \alpha =0.6 \} $ for Indian Pines, $ \{ m=1000, k=5, \alpha =0.3 \} $ for Pavia University, $ \{ m=1000, k=5, \alpha =0.8 \} $ for Salinas. For FSCS, the window size $ w $, the anchors number $ m $, neighbors number $ k $ and balance parameter $ \alpha $ for three datasets are $ \{ w=11, m=500, k=5, \alpha =0.2 \} $, $ \{ w=7, m=1000, k=5, \alpha =0.1 \} $ and $ \{ w=15, m=1000, k=5, \alpha =0.2 \} $. For the proposed SSCAG, we choose four different window scales, i.e., $ 3\times 3, 7\times 7, 11\times 11 $ and $ 15\times 15 $, to preprocess every dataset. The corresponding parameters are set as $ \{ m=500, k=5, \alpha =0.6 \} $, $ \{ m=1000, k=5, \alpha =0.4 \} $,  $ \{ m=1000, k=5, \alpha =0.7 \} $ for three HSI datasets, respectively.

\begin{figure*}[htbp]
	\centering
	\includegraphics[width=18cm,height=4cm]{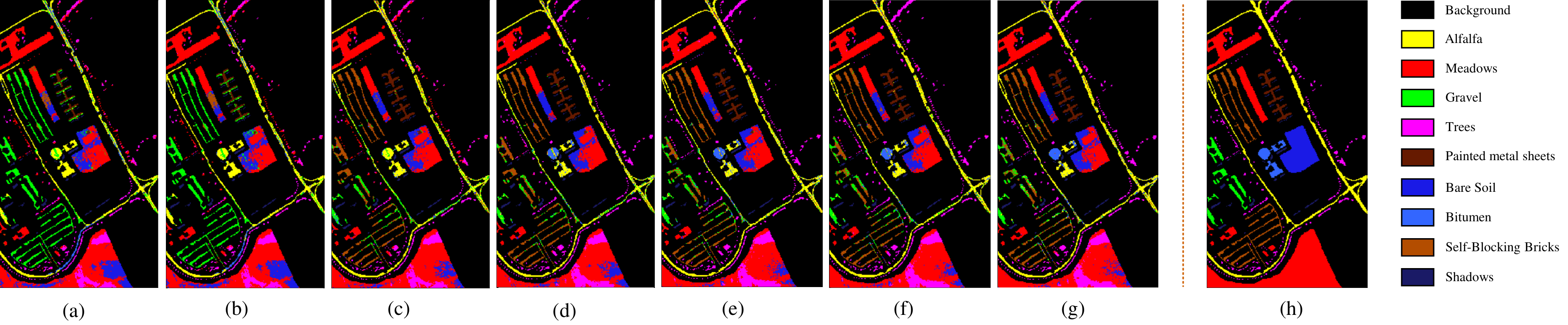}
	\caption{ Clustering maps and ground truth of Pavia University. (a)  $ k $-means. (b) FCM. (c) FCM\_S1. (d) SGCNR. (e) FSCAG. (f) FSCS. (g) SSCAG. (h) Ground truth.}\label{fig:paviau}
\end{figure*}
\begin{table*}[htbp] 
	\renewcommand{\arraystretch}{1.5}
	\caption{Quantitative metrics of different comparison methods and the proposed SSCAG on Pavia University dataset. The optimal value of each row is highlighted in bold.}
	\centering
	\label{table:paviau}
	\centering
	\begin{tabular}{|p{3.6cm}<{\centering}||p{1.1cm}<{\centering}|p{1.1cm}<{\centering}|p{1.1cm}<{\centering}|p{1.1cm}<{\centering}|p{1.1cm}<{\centering}|p{1.1cm}<{\centering}|p{1.1cm}<{\centering}|}
		\hline
		Class     &  $ k $-means& FCM & FCM\_S1 & SGCNR & FSCAG  & FSCS   & SSCAG \\ \hline
		\hline
		Asphalt	  &0.7431	 &0.7525	&0.6827	   &0.6289	  &0.8411	 &\bf0.8598	&0.8378    \\ \hline
		Meadows	  &0.6403	 &0.6395	&0.6791	   &0.7521	  &0.8141	 &0.8090	&\bf0.8245 \\ \hline
		Gravel	  &0.9827    &\bf0.9877	&0.4369	   &0.2857	  &0.5331	 &0.5089	&0.5440    \\ \hline  
		Trees	  &0.6363	 &0.6220    &0.6152	   &0.7892 	  &0.7116	 &0.7313	&\bf0.7908 \\ \hline
		Painted metal sheets &0.5424 &0.5438	&0.9865	   &0.9728	  &\bf0.9956 &0.9548	&0.9738    \\ \hline
		Bare Soil	  &0.3019	 &0.2755	&0.3106	   &0.3897	  &0.3570	 &0.4007	&\bf0.4461 \\ \hline
		Bitumen	  &0	     &0	        &0	       &0.2915	  &0.5708	 &0.5774	&\bf0.5813 \\ \hline
		Self-Blocking Bricks &0.0014 &0.0039	&0.6535	   &\bf0.8621 &0.6996	 &0.6648	&0.7451    \\ \hline
		Shadows	  &0.9978	 &0.9975	&0.9969    &0.9980	  &\bf1.0000 &\bf1.0000	&\bf1.0000 \\ \hline
		\hline
		\bf OA	  &0.5340	 &0.5341	&0.5746	   &0.6275	  &0.7306	 &0.7089	&\bf0.7504 \\ \hline
		\bf AA	  &0.5384	 &0.5358	&0.5957	   &0.6633	  &0.7248	 &0.7230	&\bf0.7493 \\ \hline
		\bf Kappa &0.4336	 &0.4334	&0.4546	   &0.5751	  &0.6497	 &0.6324	&\bf0.6747 \\ \hline	
	\end{tabular}
\end{table*}

\subsubsection{Clustering Performance Comparison} In this section, experiments are conducted on three HSI datasets. The parameter analysis of the proposed approach is also discussed.  

\textit{a) Performance on Indian Pines dataset:} The quantitative results of the methods are given in Table \ref{table:indian}. It can be revealed that SSCAG obtains better results than competitors from the perspective of OA, AA and Kappa, and gains the higher accuracy in most classes. Moreover, the AG-based methods (SGCNR, FSCAG, FSCS and SSCAG) outperform the centroid-based clustering methods ($ k $-means, FCM and FCM\_S1) as shown Table \ref{table:indian}. Among these competitors, $ k $-means and FCM are traditional clustering approaches which only focuses on the spectral information. The other algorithms (FCM\_S1, FSCAG, FSCS and SSCAG) combine spectral feature and spatial structure to enforce HSI clustering. Corresponding to the Table \ref{table:indian}, the quantitative accuracy of $ k $-means and FCM is lower (i.e., OAs reduce by at least 5\%) than the accuracy of other algorithms. This reveals that incorporating spatial information is beneficial to HSI clustering. Furthermore, compared with FSCS, the proposed SSCAG improves over 10\% in terms of OA. Because the WMF with multiscale windows outperforms the optimal results in single scale, which reaches higher OAs by complementary information in different scales.

To easily observe the classification of each class, we set the background of clustering maps as black color. Then the final clustering maps obtained with each algorithm are also visualized in Fig. \ref{fig:indian}. The proposed SSCAG obtains much smoother clustering maps than other methods. Especially, SSCAG has a superior performance than SGCNR and FSCAG, which indicates that WMF-based spatial preprocessing significantly provides an accurate description of HSI homogeneous areas. Moreover, the preprocessing increases OAs by enhancing the similarity and consistency of adjacent pixels. It reflects that multiscale WMF can provide more precise information for distance metric to construct an effective spatial-spectral adjacency graph in clustering procedure.

\textit{b) Performance on Pavia University dataset:} To observe the experimental results visually and quantitatively, Fig. \ref{fig:paviau} and Table \ref{table:paviau} show the visual clustering maps and the corresponding classification accuracy, respectively. As shown in Table \ref{table:paviau}, the AG-based clustering methods give a superior results than the classic centroid-based methods, especially for the Bitumen class. After incorporating the spatial characteristics into the anchor graph model, FSCAG, FSCS and SSCAG yield better accuracy in almost all classes compared with SGCNR, particularly for the Shadows class. This indicates that the combination of spatial and spectral information is beneficial to improve HSI clustering performance. The proposed SSCAG obtains the best results for 5 classes, except for the Gravel class, which also achieves better accuracy for other 3 classes. The OA and kappa coefficient of SSCAG achieves to 0.7504 and 0.6747, respectively. It illustrates that SSCAG is an effective and superior algorithm for HSI clustering. As is shown in Fig. \ref{fig:paviau}, SSCAG gains more smoother clustering map than other methods, which is consistent with the results in Table \ref{table:paviau}.  

\begin{figure*}[htbp]\centering
	\includegraphics[width=18cm,height=4.2cm]{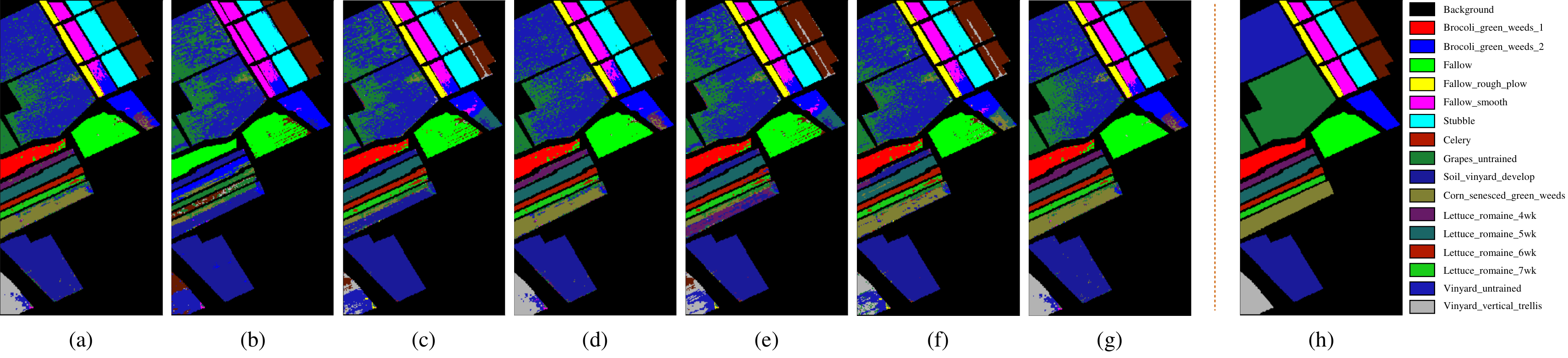}
	\caption{ Clustering maps and ground truth of Salinas. (a) $ k $-means. (b) FCM. (c) FCM\_S1. (d) SGCNR. (e) FSCAG. (f) FSCS. (g) SSCAG. (h) Ground truth.}\label{fig:salinas}
\end{figure*}

\textit{c) Performance on Salinas dataset:} To visualize the experimental results, the clustering maps are illustrated in Fig. \ref{fig:salinas}. The quantitative accuracy of comparison algorithms and SSCAG are exhibited in Table \ref{table:salinas}. According to the visual results, it reveals that the pixels in the Grapes\_untrained class and Vinyard\_untrained class regions are most wrongly assigned. That is because the Grapes\_untrained and Vinyard\_untrained have highly similar feature information, and their spectral curves are very close. From Fig. \ref{fig:salinas} (g), the Lettuce\_romaine\_7wk class is discriminated well, and its accuracy reaches to 0.9042 in Table \ref{table:salinas}, which is higher than that of other methods. As shown in Table \ref{table:salinas}, the results indicate that the proposed SSCAG yields the best classification accuracies, especially the OA and Kappa coefficient are 0.7419 and 0.6993 respectively. Furthermore, it can be observed that the Lettuce\_4wk class is effectively assigned by SSCAG, in contrast, the recognition accuracy of other algorithms except SGCNR is close to 0. This phenomenon also justifies the effectiveness of SSCAG.
  
\begin{table*}[htbp] 
	\renewcommand{\arraystretch}{1.5}
	\caption{Quantitative metrics of different comparison methods and the proposed SSCAG on Salinas dataset. The optimal value of each row is highlighted in bold.}
	\centering
	\label{table:salinas}
	\centering
	\begin{tabular}{|p{3.6cm}<{\centering}||p{1.1cm}<{\centering}|p{1.1cm}<{\centering}|p{1.1cm}<{\centering}|p{1.1cm}<{\centering}|p{1.1cm}<{\centering}|p{1.1cm}<{\centering}|p{1.1cm}<{\centering}|}
		\hline
		Class     &  $ k $-means& FCM & FCM\_S1 & SGCNR & FSCAG  & FSCS   & SSCAG \\ \hline
		\hline
		Brocoli\_green\_weeds\_1	&0.8677	&0	       &\bf0.9826&0.9501	&0.9474	&0.9300	&0.9204   \\ \hline
		Brocoli\_green\_weeds\_2	&0.4556	&\bf0.9010&0.4367	&0.5203 	&0.3695	&0.4610	&0.5678   \\ \hline
		Fallow    	        &0.7586	&0.7070	   &0.7071	&0.7825 	&0.7125	&0.7376	&\bf0.8820\\ \hline
		Fallow\_rough\_plow	    &0.9484	&0  	   &\bf0.9819&0.9241	&0.9579	&0.9507	&0.9087   \\ \hline
		Fallow\_smooth	        &0.7788	&0.8545	   &0.8972	&0.7088	    &0.8496	&0.8578	&\bf0.8985\\ \hline
		Stubble	            &0.9649	&0.9465	   &0.9545	&0.959	    &0.9587	&0.9518	&\bf0.9790\\ \hline
		Celery	            &0.9042	&\bf0.9863 &0.9055	&0.9388	    &0.8847	&0.8860	&0.9117   \\ \hline
		Grapes\_untrained	        &0.1820	&0.3780	   &0.4133	&0.3415	    &0.4486	&0.3817	&\bf0.4626\\ \hline
		Soil\_vinyard\_develop	    &0.8001	&\bf0.9713 &0.9164	&0.7917	    &0.8710	&0.8538	&0.8336   \\ \hline
		Corn\_senesced\_green\_weeds&0.5245	&0.3785	   &0.2739	&0.5023	    &0.3239	&0.5946	&\bf0.6199\\ \hline
		Lettuce\_romaine\_4wk	    &0.2849	&0.0024	   &0	    &0.6526     &0.1083	&0.2314	&\bf0.8968\\ \hline
		Lettuce\_romaine\_5wk	    &0.8789	&0.2229	   &0.9889	&\bf0.9953	&0.9805	&0.8954	&0.9525   \\ \hline
		Lettuce\_romaine\_6wk	    &0.9513	&0.1069	   &0.9782	&0.9536	    &0.7888	&\bf0.9847&0.9480 \\ \hline
		Lettuce\_romaine\_7wk	    &0.4598	&0.5879    &0.8457	&0.8501	    &0.8557	&0.8944	&\bf0.9042\\ \hline
		Vinyard\_untrained	    &0.6340	&0.5065	   &0.6295	&\bf0.6982	&0.5506	&0.5631	&0.6385   \\ \hline
		Vinyard\_vertical\_trellis	&0.4186	&0.0682	   &0.2561	&0.5058	    &0.4161	&0.4802	&\bf0.5825\\ \hline
		\hline
		\bf OA	            &0.6438	&0.4623	   &0.6754	&0.7208	    &0.6569	&0.7053	&\bf0.7419\\ \hline
		\bf AA	            &0.6758	&0.4761	   &0.6980	&0.7547	    &0.6890	&0.7284	&\bf0.8067\\ \hline
		\bf Kappa           &0.6214	&0.4038	   &0.5907	&0.6771	    &0.6216	&0.6676	&\bf0.6993\\ \hline		
	\end{tabular}
\end{table*}

\begin{figure*}[htbp]
	\centering
	\includegraphics[width=16cm,height=4cm]{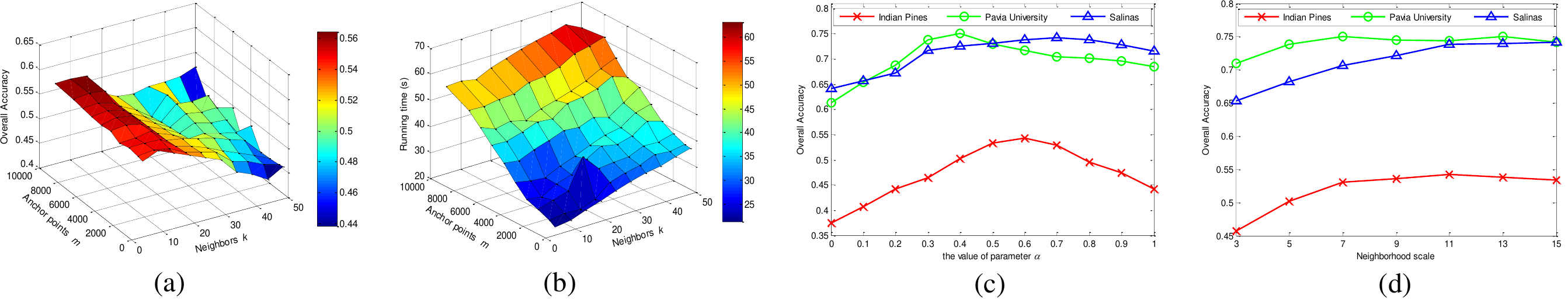}
	\caption{Parameter analysis in different situations. (a) OAs of SSCAG on Indian Pines dataset under different $ m $ and $ k $. (b) Running time of SSCAG on Indian Pines dataset under different $ m $ and $ k $. (c) OAs of SSCAG on three datasets with different values of $ \alpha $. (d) Effects of the scale $ w $ on datasets.}
	\label{fig:parameters}
\end{figure*}

\textit{d) Parameter Analysis:} The proposed SSCAG contains four parameters that need to be set and tuned, i.e. $ m $, $ w $, $ k $ and $ \alpha $. As mentioned in Section \ref{SSCAG}, the computational complexity mainly depends on the size of parameters $ m $ and $ w $. The parameter $ k $ mainly controls the sparsity of matrix $\mathbf{Z}$,  which affects the computational complexity little. Concretely, we implement the sensitivity experiments of $ m $ and $ k $ on Indian Pines dataset.  

According to Fig. \ref{fig:parameters} (a), the fluctuation range of OAs is small. We can see that the OAs of the proposed SSCAG are relatively stable with changing the parameters $ m $ and $ k $. As shown in Fig. \ref{fig:parameters} (b), the running time is mainly related to $ m $, while the influence of $ k $ is little. Hence, to speed up SSCAG algorithm and ensure good accuracy, we choose $ \{ m=500, k=5 \} $ for the Indians Pines dataset. Similarly, we select $ \{ m=1000, k=5 \}$ for the other large datasets. 

Fig. \ref{fig:parameters} (c) illustrates the OAs of three HSI datasets with varying $ \alpha $. $ \alpha $ is an important parameter in SSCAG, which is used to adjust the tradeoff between spectral feature and spatial-spectral feature during the adjacent graph learning. From Fig. \ref{fig:parameters} (c), the result demonstrates that the optimal value of $ \alpha $ is 0.6, 0.4 and 0.7 for Indian Pines, Pavia University and Salinas datasets, respectively. Fig. \ref{fig:parameters} (d) exhibits the OAs of three datasets with seven different neighborhood scales. The parameter $ w $ is the window scale of WMF that affects the result of preprocessing. Here, we adopt seven different scales (i.e., $ 3 \times 3 $, $ 5 \times 5 $,...,$ 15 \times 15 $) to investigate its impact. As shown in Fig. \ref{fig:parameters} (d), the best OAs for Indian Pines, Pavia University and Salinas are obtained, when the window scale is 11, 7 and 15 respectively. This indicates that different HSI datasets involve different spatial structures, even the same dataset may include small and large homogeneous regions simultaneously. So it is difficult to determine a best scale in advance. Therefore, we adopt four scales ($ 3\times 3$, $7\times 7$, $ 11\times 11 $ and $ 15\times 15 $) in the multiscale WMF framework, which not only exempts from the scale selection but also provides multiple views of a local homogeneous region. By doing this, it obtains the complementary information and improve the final clustering performance.

\begin{table*}[htbp] 
	\renewcommand{\arraystretch}{1.5}
	\caption{Running time of different methods on three HSI datasets}
	\centering
	\label{table:time}
	\centering
	\begin{tabular}{|p{3.6cm}<{\centering}||p{1.1cm}<{\centering}|p{1.1cm}<{\centering}|p{1.1cm}<{\centering}|p{1.1cm}<{\centering}|p{1.1cm}<{\centering}|p{1.1cm}<{\centering}|p{1.1cm}<{\centering}|}
		\hline
		Dataset & $ k $-means& FCM  & FCM\_S1  & SGCNR  & FSCAG  & FSCS & SSCAG \\ \hline
		\hline
		Indian Pines  &5.6s  &7.1s	 &9.6s   &10.6s   &10.1s  &9.9s   &21.3s \\ \hline
		Pavia University&26.3s &36.7s &49.3s	 &121.7s  &137.5s &191s	  &130.8s\\ \hline
		Salinas    &20.1s &27.4s &37.7s  &45.8s   &26s	  &25.7   &61.1s  \\ \hline  	
	\end{tabular}
\end{table*}   

\subsubsection{Running Time Comparison}
We display the running time to quantitatively compare the complexity of all algorithms, as shown in Table \ref{table:time}. All of the experimental results are conducted in MATLAB R2014a on a PC of Intel Core i7-9700F 3.00GHz CPU with 16 GB RAM. As shown in Table \ref{table:time}, $ k $-means consumes the least running time, but its accuracy is generally not high comparing with other algorithms (see Table \ref{table:indian}, \ref{table:paviau} and \ref{table:salinas}). From Table \ref{table:time}, the proposed SSCAG method is slower than the other algorithms on the HSI datasets except Pavia University dataset. However, according to Table \ref{table:indian}, \ref{table:paviau} and \ref{table:salinas}, SSCAG provides best results than other methods, especially for AAs, which are 6\%, 2.5\% and 5\% higher than the second best results on Indian Pines, Pavia University and Salinas datasets respectively. Therefore, the slight increase in running time is acceptable for the improvement in clustering performance. 

\section{Conclusion}
In this paper, a novel spatial-spectral clustering with anchor graph (SSCAG) is proposed to efficiently cluster HSI data. Firstly, the multiscale spatial WMF is utilized to enhance the local pixel consistency and distinguish the structures across different classes. To facilitate the graph construction, we select representative anchors and exploit their relationship with the points, which reduces the computational complexity of model. Secondly, a new spatial-spectral distance metric is proposed to combine the comprehensive features, which reveals the inherent properties of HSI data. Finally, the neighbors assignment strategy is used to learn the optimal adjacent graph adaptively, and SVD is performed on $ \mathbf {S}$ to get the final clustering results. Extensive experiments on three public HSI datasets verify the efficiency and effectiveness of the proposed SSCAG. In the future, we mainly focus on designing a fusion strategy with multi-graph construction to better handle the tasks of HSI clustering.

\bibliographystyle{IEEEbib}

\bibliography{ref}

\end{document}